\documentclass[letterpaper]{article} 
\usepackage{aaai2026}  
\usepackage{times}  
\usepackage{helvet}  
\usepackage{courier}  
\usepackage[hyphens]{url}  
\usepackage{graphicx} 
\usepackage{natbib}  
\usepackage{caption} 
\usepackage{algorithm}
\usepackage{algorithmic}

\usepackage{xcolor}

\usepackage{newfloat}
\usepackage{listings}
\DeclareCaptionStyle{ruled}{labelfont=normalfont,labelsep=colon,strut=off} 
\floatstyle{ruled}
\newfloat{listing}{tb}{lst}{}
\floatname{listing}{Listing}
\nocopyright

\usepackage{multirow}
\usepackage{lipsum}
\usepackage{booktabs}
\usepackage{amsthm}
\usepackage{amsmath}
\usepackage{amsfonts}
\usepackage{mathrsfs}

\newtheorem{definition}{Definition}

\def \k {\kappa}

\def \x {\mathbf{x}}
\def \A {\mathcal{A}}
\def \C {\mathcal{C}}
\def \D {\mathbf{D}}
\def \E {\mathbb{E}}

\def \O {\mathcal{O}}
\def \P {\mathcal{P}}
\def \R {\mathbb{R}}
\def \X {\mathbf{X}}

\newcommand{\tabincell}[2]{\begin{tabular}{@{}#1@{}}#2\end{tabular}}

\newcommand\blfootnote[1]{%
  \begingroup
  \renewcommand\thefootnote{}\footnote{#1}%
  \addtocounter{footnote}{-1}%
  \endgroup
}

\title{Distribution-Based Feature Attribution\\ for Explaining the Predictions of Any Classifier}
\author{
    Xinpeng Li,
    Kai Ming Ting
}
\affiliations{
    State Key Laboratory for Novel Software Technology, Nanjing University, Nanjing, China\\
    School of Artificial Intelligence, Nanjing University, Nanjing, China\\
    lixp@lamda.nju.edu.cn,
    tingkm@nju.edu.cn
}

\begin{document}

\maketitle

\begin{abstract}
The proliferation of complex, black-box AI models has intensified the need for techniques that can explain their decisions.
Feature attribution methods have become a popular solution for providing post-hoc explanations, yet the field has historically lacked a formal problem definition.
This paper addresses this gap by introducing a formal definition for the problem of feature attribution, which stipulates that explanations be supported by an underlying probability distribution represented by the given dataset.
Our analysis reveals that many existing model-agnostic methods fail to meet this criterion, while even those that do often possess other limitations.
To overcome these challenges, we propose Distributional Feature Attribution eXplanations (DFAX), a novel, model-agnostic method for feature attribution.
DFAX is the first feature attribution method to explain classifier predictions directly based on the data distribution.
We show through extensive experiments that DFAX is more effective and efficient than state-of-the-art baselines.
\end{abstract}


\blfootnote{Accepted for an oral presentation at the 40th Annual AAAI Conference on Artificial Intelligence (AAAI-26). This document is the extended version which includes the appendix.}

\section{Introduction}

Recent years have witnessed a rapid growth and widespread adoption of artificial intelligence (AI).
However, a major challenge remains: most state-of-the-art models are black-boxes which are unable to explain their own decisions.
This lack of transparency has motivated the development of explainable AI (XAI) techniques, a field dedicated to helping users understand and trust these models \cite{minh2022explainable}.

Feature attribution has emerged as a crucial XAI technique for providing post-hoc explainability by computing contribution scores, which quantify the importance of input features with respect to the output of a model \cite{arrieta2020explainable}.
These methods are broadly categorized as either model-specific or model-agnostic.
The model-specific approach leverages the knowledge concerning the model's internal structure.
For example, numerous methods have been proposed for deep neural networks (DNNs) \cite{zhu2024iterative, walker2024integrated}, and they often require access to differentiable gradients in DNN.
This reliance restricts their applicability to DNN only.
In contrast, model-agnostic methods treat a model as a black-box, generating feature attributions solely by observing and analyzing the model's input-output behavior.
This approach makes them universally applicable to any model, regardless of model types.
Major families of model-agnostic feature attribution methods include local approximation and perturbation-based approaches \cite{li2023negative, ivanovs2021perturbation}.

Despite years of research and the development of numerous methods, the field of feature attribution has, until now, lacked a formal problem definition.
This foundational gap complicates the analysis and comparison of different methods.
In this paper, we are motivated to provide a formal definition for the task of feature attribution.
Our proposed definition serves as a foundational criterion, establishing a clear standard for the analysis of specific methods.

Furthermore, our analysis of existing methods reveals that even those that satisfy our proposed problem definition often suffer from other limitations restricting their performance or practical applicability.
To address these issues, we introduce Distributional Feature Attribution eXplanations (DFAX), a novel model-agnostic method for explaining classifier predictions based on the probability distribution of a given dataset.
DFAX is designed to adhere to our formal problem definition while simultaneously overcoming the key limitations of prior approaches.

The main contributions of this work are:
\begin{itemize}
    \item We introduce a formal definition for the problem of feature attribution, which provides a key criterion for the analysis and design of feature attribution methods.
    \item With this problem definition, we conduct an analysis of existing methods, identifying their individual limitations.
    \item We propose the DFAX scheme which complies with the definition and does not have the identified limitations of prior methods.
    To the best of our knowledge, DFAX is the first explainer to approach feature attribution by directly leveraging the underlying data distribution.
    \item Through extensive quantitative and qualitative experiments, we demonstrate the superior effectiveness and efficiency of DFAX compared to other state-of-the-art model-agnostic methods for feature attribution.
\end{itemize}

\begin{figure*}[t]
\centering
\includegraphics[width=0.87\textwidth]{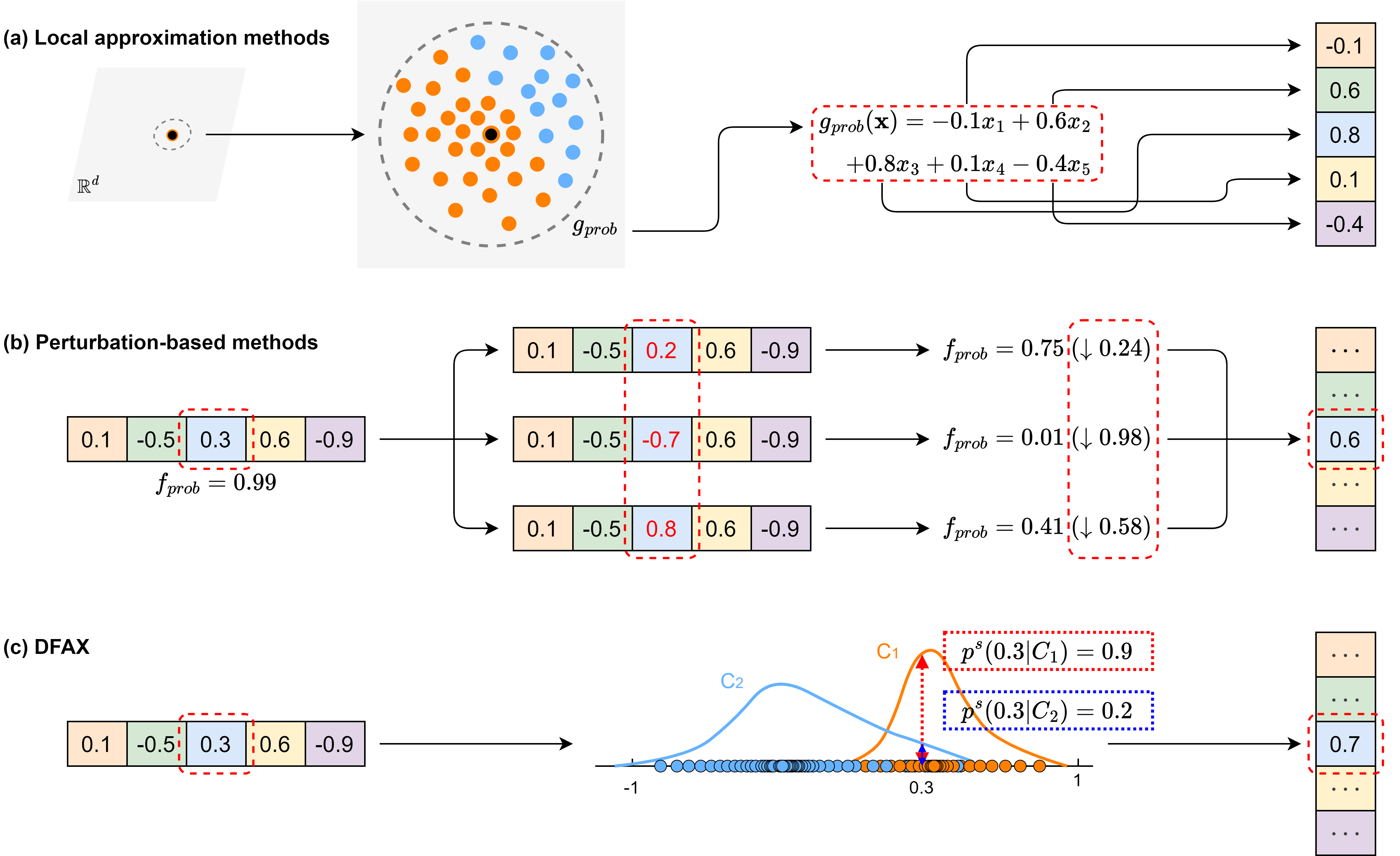}
\caption{
Illustrations of two existing families of methods, shown in (a) \& (b);
and the proposed DFAX, shown in (c).
}
\label{fig:illustration}
\end{figure*}

\section{Related Work}

This section reviews the two predominant families of model-agnostic feature attribution methods, followed by an overview of kernel density estimation, a core component of our proposed DFAX method.

\subsection{Local Approximation Methods}

Local approximation methods, as illustrated in Figure~\ref{fig:illustration}(a), operate by first defining a neighborhood around the target instance and then fitting a simple surrogate model to the classifier's predictions within this local region.
Feature attributions are then derived from this local surrogate model.

A prominent example is LIME \cite{ribeiro2016should}, which generates its neighborhood by creating synthetic points around the target instance via random sampling before fitting an explanatory linear model.
DLIME \cite{zafar2021deterministic}, a deterministic version of LIME, leverages agglomerative hierarchical clustering on the training data to define the neighborhood.
Similarly, MAPLE \cite{plumb2018model} and SLISE \cite{bjorklund2023explaining} select neighboring points from the given dataset instead of generating synthetic points as done in LIME.
MAPLE weights these neighbors to produce more faithful local explanations, while SLISE fits a sparse, locally linear regression model whose coefficients serve as feature attributions.
Focusing on improving stability and unidirectionality, LINEX \cite{dhurandhar2023locally} minimizes sensitivity to perturbations in a way inspired by the invariant risk minimization (IRM) principle.
The specific implementation of LINEX can vary, as the local environments for IRM may be created differently depending on the application scenario.

\subsection{Perturbation-Based Methods}

As depicted in Figure~\ref{fig:illustration}(b), perturbation-based methods operate by perturbing a feature's value and measuring the resulting degradation in the classifier's performance (\textit{e.g.}, the decrease in the predicted probability for the target class).

A famous method in this category is SHAP \cite{lundberg2017unified}, which is grounded in cooperative game theory and calculates feature importance using Shapley values \cite{shapley1953value}, the marginal contribution of each feature to the prediction for the target instance.
The practical implementation of SHAP varies based on how this contribution is estimated.
A theoretically pure approach, based on Shapley regression values \cite{lipovetsky2001analysis}, requires retraining the classifier on all possible subsets of features.
Common implementations instead use Shapley sampling values \cite{strumbelj2014explaining}.
These methods avoid retraining by using the original classifier trained with all feature present, and observing its performance changes on perturbed instances created by combining feature-values from the target instance and a background dataset.

Another perturbation-based method is PFI \cite{fisher2019all}, a model-agnostic version of the original PFI \cite{breiman2001random}.
PFI measures the importance of a feature by the expected loss in classifier performance after permuting the feature with values across the entire dataset.

\subsection{Kernel Density Estimation}

A kernel density estimator (KDE) is a non-parametric method for estimating the density/probability of each point in a given dataset \cite{rosenblatt1956remarks, ting2021isolation}.
Given a dataset $\X\subset\R^d$, the KDE at any point $\x^*\in\R^d$ is defined as:
\begin{equation*}
    K(\x^*|\X)=\frac1{|\X|}\sum_{\x\in\X}\k(\x^*,\x)
\end{equation*}
where $\k$ is a point kernel.
Different choices of $\k$ yield different estimators.
For example, a Gaussian kernel results in the Gaussian Kernel Density Estimator (GKDE), while the Isolation Kernel \cite{ting2018isolation} produces Isolation Kernel Density Estimator (IKDE) \cite{ting2021isolation}.

A fast alternative to GKDE in outlying aspects mining tasks \cite{wells2019new} is SiNNE \cite{samariya2020new}, which is a simplified version of iNNE \cite{bandaragoda2014efficient}.
While originally designed to estimate anomaly scores, SiNNE can be used as an efficient substitute for KDE in our feature attribution task.

The KDE computation can be significantly accelerated if the point kernel can be approximated as $\k(x,y)\approx\left<\varphi(x), \varphi(y)\right>$ via some technique like the Nystr{\"o}m method \cite{williams2000using}, where $\varphi$ is a finite-dimensional feature map approximating the feature map of $\k$.
With this, the KDE can be re-expressed as:
\begin{equation*}
     K(\x^*|\X)\approx\left<\varphi(\x^*), \widehat\Phi(\X)\right>
\end{equation*}
where $\widehat\Phi(\X)=\frac1{|\X|}\sum_{\x\in\X}\varphi(\x)$ is the kernel mean map of $\X$ \cite{muandet2017kernel}.
This allows any subsequent probability/density estimation to be performed in $\O(1)$ time after a one-off computation for the kernel mean map of $\X$.

\section{Problem Definition}

When the objective is to understand the model's logic on its operational data distribution, the problem of feature attribution can be formally defined as follows:

Let $\A=\{s_j\}_{j=1}^d$ be the set of features and $m$ be the total number of classes.
Let $f$ be a classifier trained on a dataset $\D\subset\R^d$, which maps inputs to either a predicted class or a predicted probability for a target class ($f\colon\R^d\mapsto[m]$ or $[0,1]$).
Let $\x^*\in\R^d$ be the target instance and $\X=\{\x_i\}_{i=1}^n\subset\R^d$ be a given dataset.
We assume $\x^*$, $\X$, and $\D$ are all independent and identically distributed (i.i.d.) samples from the same underlying probability distribution $\P$.

\begin{definition}[Feature Attribution]
\label{def:FA}
For a target instance $\x^*\sim\P$ with features $\A$ whose prediction $y^*=f(\x^*)$ is produced by classifier $f$, the task of feature attribution aims to provide an explanation as a score
$I(\x^*,s|\X)$ to each feature $s\in\A$.
This score quantifies the influence of the specific feature-value, $\x_s^*$, on the classifier $f$ to produce the prediction $y^*$, where a higher score indicates a greater influence towards this prediction.
The explanatory model, $I(\cdot|\X)$, must be built directly from the dataset $\X$, which reflects the underlying distribution $\P$, and the score $I(\x^*,s|\X)$ is valid if and only if it is supported by $\P$.
\end{definition}

A crucial tenet of this definition is the role of the dataset $\X$, which serves as an empirical representation of the underlying distribution $\P$.
Any modification to $\X$ that changes the underlying distribution, in the process of building the explanatory model $I(\cdot|\X)$, invalidates its feature attribution.
This is because building the explanatory model using synthetic or out-of-distribution (OOD) instances produces explanations based on a distribution where the model's behavior is irrelevant or inapplicable.
In a nutshell, the key criterion of Definition~\ref{def:FA} is that
\emph{the explanatory model $I(\cdot|\X)$ and its explanation $I(\x^*,s|\X)$ must be supported by the distribution $\P$ which is represented by the unmodified dataset $\X$ (no OOD instance is used for generating feature attributions).}

\section{Analyses of Existing Methods}

Definition~\ref{def:FA} provides the criterion for assessing any feature attribution method.
In this section, we analyze the two predominant families of model-agnostic methods.

(a) \textbf{Local approximation methods}.
Some of these methods satisfy Definition~\ref{def:FA}.
For example, given the training dataset $\D=\X$ for the classifier $f$, DLIME \cite{zafar2021deterministic} selects a neighboring cluster around the target instance from $\X$, upon which it fits a linear regression model $LR(\x)=\omega\x+b$.
$LR$ works as the explanatory model $I$ in Definition~\ref{def:FA}.
For each feature $s_i\in\A$ of the target instance $\x^*$, the coefficient $\omega_i$ in $LR$ corresponds to the score $I(\x^*,s_i|\X)$ stated in the definition.
Note that while DLIME utilizes the unmodified $\X$ to build the explanatory model, thus satisfying Definition~\ref{def:FA}, it only uses a small subset of $\X$ which corresponds to a local region that DLIME focuses on to build $LR$, due to its reliance on fitting a simple surrogate model in a selected local neighborhood.
This is an inherent limitation of local approximation methods.

Other methods in this family, such as LIME \cite{ribeiro2016should}, not only share this limitation of a local focus but also fail to satisfy the criterion of Definition~\ref{def:FA}.
This is because their explanatory models are fitted to a synthetic neighborhood generated via random perturbation, independent of distribution $\P$.

(b) \textbf{Perturbation-based methods}.
Some of these methods, such as an implementation of SHAP \cite{lundberg2017unified} that calculates the Shapley regression values \cite{lipovetsky2001analysis}, comply with Definition~\ref{def:FA}.
The score it produces is given by:
\begin{equation}
\label{eq:SHAP}
\begin{aligned}
    I(\x^*,s|\X)&=
    \sum_{S\subseteq\A\setminus\{s\}}M_S\left    (F_{S\cup\{s\}}(\x^*)-F_{S}(\x^*)\right)\\
    \text{where}\quad M_S&=\frac{|S|!(|\A|-|S|-1)!}{|\A|!}
\end{aligned}
\end{equation}
where $n!$ is the factorial of $n$, and $F_{S}$ is the estimated probability function of the classifier $f$ retrained using only a subset of features $S\subseteq\A$, with all other features withheld.

In this case, $\D=\X$ in the subspace defined by the feature subset $S$ is used for building the explanatory model $I$, satisfying the criterion of being supported by $\P$ in Definition~\ref{def:FA}.

However, calculating Shapley regression values is computationally infeasible.
Practical implementations commonly use Shapley sampling values \cite{strumbelj2014explaining}, which introduce two key approximations.
First, instead of iterating through the entire power set of $\A\setminus\{s\}$ with $2^{d-1}$ feature subsets, Equation~\ref{eq:SHAP} is approximated by sampling a tractable number of subsets.
Second, to avoid repeatedly retraining the classifier, $F_{S}(\x^*)$ is replaced by $\E\left[f_{prob}(\tilde\x_S)\ |\ \tilde\x_S\in\tilde{\X}_S\right]$ which is the expected output of the classifier $f_{prob}$.
As $f_{prob}$ is trained on $\D$ with all features, the dataset $\X$ is modified for the feature subset $S$ as follows: 
$\tilde{\X}_S$ contains instances originally the same as $\X$, except that for each instance $\tilde\x_S\in\tilde{\X}_S$, its feature-values in the subset $S$ are replaced with those in the target instance $\x^*$, while the values for the remaining features are kept unchanged.

While enhancing efficiency, this approximation method of SHAP does not satisfy Definition~\ref{def:FA}, because it modifies $\X$ into $\tilde\X_S$, mixing feature-values from $\x^*$ and the instances in $\X$.
These synthetic instances invalidate the key criterion of feature attribution stated in Definition~\ref{def:FA}, \textit{i.e.}, they are not i.i.d. samples from the distribution $\P$ for which the classifier was trained.
Consequently, the explanation is inappropriate as it is not supported by $\P$.
This is a common oversight for most methods in this family, a result of designing a method without knowing the problem definition.

Note that while methods like LIME and Shapley sampling values do not satisfy Definition~\ref{def:FA}, they address a different objective: understanding model behavior across the entire input space, rather than on its operational data distribution \cite{chen2020true}.
However, this objective has limited practical applications.
For most real-world XAI purposes, the primary goal is to understand model behavior specifically on the data distribution it was trained and qualified to operate on \cite{freiesleben2023dear}.

Definition~\ref{def:FA} provides the criterion for assessing whether a method has any compliance issues. 
Our analysis above shows that while some existing methods satisfy the criterion stated in Definition~\ref{def:FA}, others do not.
A serious oversight of existing methods is the misuse of $\X$, creating synthetic instances that violate the underlying distribution $\P$.
This fundamentally conflicts with the objective of understanding a model's logic on its operational data distribution.
Another limitation of many methods is the inability to fully utilize $\X$.
The above analyses reveal the importance of Definition~\ref{def:FA}, which serves as a guide in designing a reliable method that complies with the objective and makes the full use of $\X$.

\section{Proposed Method: DFAX}

With the above-mentioned limitations of existing methods, we are motivated to develop a better approach that not only solves the problem of feature attribution defined in Definition~\ref{def:FA}, but also overcomes the identified limitations.
A simple yet effective solution is via a distributional approach.

Let the predicted class for the target instance be $y^*=f(\x^*)$, and the corresponding predictions for $\X$ be $\{y_i\}_{i=1}^n=\{f(\x_i)\}_{i=1}^n$.
In the subspace defined by any feature $s\in\A$, the distribution, \textit{i.e.}, the conditional probability, for a subset of classes $\C\subseteq[m]$, denoted as $p^s(\cdot|\C)$, can be estimated from the subset of points $\X_{\C}=\{\x_i\ |\ \x_i\in\X\text{ and }y_i\in\C\}$.
Our proposed Distributional Feature Attribution eXplanations (DFAX) method is then defined as:

\begin{definition}[DFAX]
\label{def:DFAX}
    Given the target instance $\x^*$ and feature $s\in\A$, DFAX computes the score as the difference between the conditional probability of $\x^*$ given the target class and that given all the other classes:
    \begin{equation*}
    \begin{aligned}
        I(\x^*,s|\X)&=p^s(\x^*\ |\ \{y^*\})-p^s(\x^*\ |\ [m]\setminus \{y^*\})\\
        &=K^s\left(\x^*|\X_{\{y^*\}}\right)-K^s\left(\x^*|\X\setminus\X_{\{y^*\}}\right)
    \end{aligned}
    \end{equation*}
    where the probability is computed using a KDE $K^s$ in the one-dimensional subspace defined by feature $s$.
\end{definition}

In contrast to the two major families of model-agnostic feature attribution methods, DFAX operates directly from a distributional perspective, as illustrated in Figure~\ref{fig:illustration}(c).
By measuring the probabilities based on the feature $s$, conditional on some classes, DFAX quantifies the extent to which the feature-value of the target instance $\x^*$ characterizes the data points belonging to the target class in $\X$, while it does not characterize points from all other classes at the same time.
To the best of our knowledge, DFAX is the first feature attribution approach based on this principle.

DFAX satisfies Definition~\ref{def:FA} as the unmodified $\X$ from distribution $\P$ is used to estimate conditional probability.
The role of $\X$ in DFAX is similar to that of the training set used in the k-nearest neighbors algorithm.
Following the lazy learning approach, DFAX defers the computation of probability until the target instance is provided, rather than learning an explicit explanatory model beforehand.

Compared with existing methods that comply with Definition~\ref{def:FA}, DFAX makes full use of the global information contained in the entire dataset $\X$, rather than limiting itself to a local region that corresponds only to a subset of $\X$.
Moreover, DFAX allows for significant computational acceleration in practice.
If the kernel used for density estimation has (or can be approximated by) a finite-dimensional feature map, the kernel mean map of $\X$ can be pre-computed before the target instance is provided.
This one-off pre-computation dramatically speeds up the attribution process, especially when there are multiple target instances to explain.

DFAX has several other desirable characteristics.
First, its feature attribution process is fully decoupled from the classifier.
DFAX operates solely on $\X$ and its pre-computed predictions, eliminating the need for further queries to the classifier $f$, which is ideal for scenarios where the classifier is expensive to query or unavailable due to privacy or proprietary concerns.
By substituting predictions with ground-truth labels, this decoupling also enables DFAX to explain the data's inherent class structure, independent of any specific classifier.
Second, DFAX utilizes global distributional information, generating attributions based on characteristic properties of the target class and non-target classes.
This is distinct from local methods that are often sensitive to neighborhood selection \cite{visani2020optilime}.

\section{Experiments}

In this section, we demonstrate the effectiveness and efficiency of the proposed DFAX method through both quantitative and qualitative evaluations.

\begin{table}[t]
\centering
\begin{tabular}{l|r|r|r|r}
\toprule
Datasets & \#inst & \#feat & \#cls & Classifier, acc\\
\midrule
Diabetes & 520 & 16 & 2 & RF, .98\\
HER2st & 527 & 314 & 6 & RF, .73\\
Rice & 3,810 & 7 & 2 & LR, .91\\
Waveform & 5,000 & 40 & 3 & LR, .86\\
Bankruptcy & 6,819 & 95 & 2 & AB, .97\\
RottenTomatoes & 10,662 & 300 & 2 & NB, .76\\
Pendigits & 10,992 & 16 & 10 & SVM, .99\\
DryBean & 13,611 & 16 & 7 & SVM, .98\\
MNIST & 70,000 & 784 & 10 & MLP, .99\\
FMNIST & 70,000 & 784 & 10 & ResNet, .94\\
\bottomrule
\end{tabular}
\caption{
Datasets and classifiers used in the experiments, along with the characteristics of the datasets and testing accuracies of the classifiers.
\#inst stands for the number of instances, feat for features, cls for classes, RF for Random Forest, LR for Logistic Regression, AB for AdaBoost, NB for Naive Bayes, SVM for Support Vector Machine, MLP for Multi-Layer Perceptron, and ResNet for Residual Network.
}
\label{tab:datasets}
\end{table}

\subsection{Experimental Setup}

All experiments were conducted on a server equipped with an AMD EPYC 7742 CPU, an NVIDIA RTX A6000 GPU, and 1TB of memory.
The system was running Ubuntu 20.04.2 (Linux kernel 5.4.0) with Python 3.9.12.
Unless specified otherwise, we employ GKDE as the default KDE in our implementation of DFAX.

\subsubsection{Datasets.}

The experiments are conducted on ten real-world datasets covering diverse modalities including tabular, text, and image, many of which have been utilized in prior studies \cite{asuncion2007uci, lecun2002gradient, xiao2017fashion, andersson2020spatial, pang2005seeing}.
All features within each dataset are standardized.
We randomly select 100 samples from each dataset as the testing set, with the remaining samples forming the training set.
The training set $\D$ is used as $\X$ when required by a feature attribution method, while the testing set also serves as the set of target instances whose attributions are sought in the quantitative experiments.

\subsubsection{Classifiers.}

For each dataset, we train a classifier on the training set and use it for prediction of the target instances in the testing set.
A wide range of classifiers, such as support vector machine with radial basis function kernel, ensemble methods, and neural networks, are employed in this process \cite{breiman2001random, freund1995desicion, he2016deep}.
Table~\ref{tab:datasets} provides the details of the datasets and classifiers.

\subsubsection{Baseline methods.}

We compare DFAX against five state-of-the-art model-agnostic feature attribution methods:
LINEX \cite{dhurandhar2023locally}, SLISE \cite{bjorklund2023explaining}, SHAP \cite{lundberg2017unified}, MAPLE \cite{plumb2018model}, and DLIME \cite{zafar2021deterministic}.
These selected baselines represent the two predominant families of methods: local approximation and perturbation-based.
SLISE, MAPLE, and DLIME comply with the criterion of Definition~\ref{def:FA}, while the implementations of LINEX and SHAP we use here do not.
It should be noted that while variants of LINEX and SHAP exist that also satisfy this definition, they have other limitations that restrict their general use.
Therefore, to ensure a practical comparison, we evaluate the most widely applicable implementations of LINEX and SHAP.
In addition, a random baseline is incorporated as a sanity check in the quantitative evaluation.
The hyperparameter specifications for all methods and datasets are provided in Appendix~\ref{app:params}.

\begin{table*}[t]
\centering
\begin{tabular}{c|c|c|c|c|c|c|c|c|c}
\toprule
Metrics & Datasets & DFAX$_\text{G}$ & DFAX$_\text{S}$ & LINEX & SLISE & SHAP & MAPLE & DLIME & Random\\
\midrule
\multirow{12}{*}{\tabincell{c}{Deletion\\Score $\downarrow$}} & Diabetes & .5442 & \textbf{.5420} & .6918 & .7058 & .6634 & .6744 & \underline{.5422} & .7182\\
& HER2st & \textbf{.1247} & \underline{.2395} & .4393 & .4358 & .3911 & .4725 & .4560 & .4672\\
& Rice & \underline{.6387} & \textbf{.6321} & .6849 & .6859 & .7050 & .8203 & .7458 & .7224\\
& Waveform & \textbf{.2621} & \underline{.2650} & .4544 & .4559 & .4905 & .5814 & .4335 & .6020\\
& Bankruptcy & \textbf{.5944} & \underline{.6129} & .6166 & .6164 & .6765 & .6169 & .6277 & .6312\\
& RottenTomatoes & \underline{.1213} & \textbf{.0682} & .6120 & .7395 & .5484 & .7397 & .3138 & .7034\\
& Pendigits & \textbf{.2529} & \underline{.2645} & .3766 & .4041 & .3651 & .3997 & .2845 & .4248\\
& DryBean & \textbf{.3148} & \underline{.3263} & .4705 & .4786 & .5397 & .4662 & .4754 & .4933\\
& MNIST & \underline{.1608} & \textbf{.1535} & .5847 & .6456 & .4779 & .5932 & .4591 & .6351\\
& FMNIST & \textbf{.2299} & \underline{.2403} & .3565 & .2894 & .3887 & .3069 & .2567 & .3112\\
\cmidrule{2-10}
& Average & \textbf{.3244} & \underline{.3344} & .5287 & .5457 & .5246 & .5671 & .4595 & .5709\\
\cmidrule{2-10}
& Avg. Ranking & \textbf{1.5} & \underline{1.6} & 4.8 & 5.5 & 5.4 & 6.1 & 4.1 & 7.0\\
\midrule
\multirow{12}{*}{\tabincell{c}{Insertion\\Score $\uparrow$}} & Diabetes & \underline{.8738} & .8727 & .7643 & .7375 & .7352 & .7789 & \textbf{.8781} & .7454\\
& HER2st & \textbf{.8119} & \underline{.6962} & .5023 & .5078 & .4648 & .4749 & .4851 & .4762\\
& Rice & \underline{.8028} & \textbf{.8136} & .7809 & .7800 & .7260 & .6459 & .7142 & .7389\\
& Waveform & \textbf{.8906} & \underline{.8820} & .7415 & .7338 & .6371 & .6364 & .7697 & .6154\\
& Bankruptcy & \textbf{.6557} & \underline{.6387} & .6354 & .6353 & .5720 & .6341 & .6238 & .6204\\
& RottenTomatoes & \textbf{.9907} & \underline{.9904} & .8714 & .7494 & .6237 & .7483 & .9550 & .7668\\
& Pendigits & \textbf{.5903} & \underline{.5524} & .4462 & .4171 & .4024 & .4146 & .5407 & .4095\\
& DryBean & \textbf{.6681} & \underline{.6570} & .5246 & .5236 & .4425 & .5323 & .5371 & .5039\\
& MNIST & \textbf{.9269} & \underline{.9212} & .6293 & .6622 & .4794 & .6137 & .7128 & .6494\\
& FMNIST & \textbf{.4975} & \underline{.4454} & .4302 & .3166 & .3797 & .3205 & .3952 & .3119\\
\cmidrule{2-10}
& Average & \textbf{.7708} & \underline{.7470} & .6326 & .6068 & .5463 & .5800 & .6612 & .5838\\
\cmidrule{2-10}
& Avg. Ranking & \textbf{1.2} & \underline{2.0} & 4.1 & 5.1 & 7.3 & 6.1 & 3.8 & 6.4\\
\bottomrule
\end{tabular}
\caption{
Comparison of different model-agnostic feature attribution methods based on deletion and insertion scores.
The results are averaged over 100 target instances $\times$ 100 random trials.
DFAX$_\text{G}$ and DFAX$_\text{S}$ employ GKDE and SiNNE as the kernel density estimator, respectively.
$\downarrow$ indicates that a lower value for the evaluation metric is better, while $\uparrow$ indicates the higher the better.
The \textbf{best} results are shown in boldface and the \underline{second-best} results are underlined.
}
\label{tab:d&i}
\end{table*}

\subsection{Quantitative Evaluation}

First, we quantitatively verify the effectiveness of DFAX through extensive comparative experiments.

\subsubsection{Evaluation metrics.}

We adopt the deletion and insertion scores \cite{petsiuk2018rise} as our evaluation metrics.
The deletion score measures the Area Under the Curve (AUC) of the classifier’s predicted probability for the target class of a given instance, as important features found by a feature attribution method are progressively masked with random values drawn from a standard normal distribution, while the insertion score does so as features are incrementally reintroduced.
These scores quantify the fidelity of the attribution to the classifier \cite{klein2024navigating}.
They have been widely adopted in the literature as established metrics for evaluating feature attribution methods \cite{walker2024integrated, li2023negative, zhu2024iterative}.

\subsubsection{Results.}

Table~\ref{tab:d&i} presents the quantitative evaluation results for different model-agnostic feature attribution methods based on the deletion and insertion scores, along with their average scores and average rankings across all datasets.

The proposed DFAX (DFAX$_\text{G}$ and DFAX$_\text{S}$ implementations) significantly outperforms other baseline methods, securing the first or second rank on nine of the ten datasets.
The only exception is the Diabetes dataset, where the performances of the top three methods are comparable.
Between the two implementations, DFAX$_\text{G}$ generally shows better results than DFAX$_\text{S}$, especially in terms of the insertion score.

Among other baseline methods evaluated, DLIME exhibits the strongest performance, followed by LINEX.
However, DFAX outperforms both of them by a large margin.
Note that the average insertion scores for both SHAP and MAPLE fall below the random baseline, thus failing the sanity check.
Some of the baseline methods, \textit{i.e.}, LINEX and SHAP, do not comply with the criterion of being supported by the distribution $\P$ stated in Definition~\ref{def:FA}.
While others (DLIME, SLISE, and MAPLE) do, they do not fully exploit the information in $\X$.

Two key factors contribute to the outstanding performance of DFAX.
First, DFAX derives feature attributions supported by the underlying data distribution $\P$ which is empirically represented by the provided dataset $\X$.
Second, DFAX makes full use of $\X$, leveraging global distributional information from the entire dataset.

\begin{table*}[t]
\centering
\begin{tabular}{|l|l|l|}
\toprule
Positive example 1 & Positive example 2 & Positive example 3\\
\midrule
\tabincell{l}{the film's \textit{real} appeal won't be to clooney\\fans or adventure buffs, but to moviegoers\\who enjoy thinking about \textbf{compelling}\\questions with no easy answers} & \tabincell{l}{a \textbf{fascinating}, bombshell \textit{documentary}\\that should shame americans$\dots$} & \tabincell{l}{$\dots$\textbf{moving} film that respects its\\\textit{audience} and its source material}\\
\bottomrule
\toprule
Negative example 1 & Negative example 2 & Negative example 3\\
\midrule
\tabincell{l}{contains the \textit{humor}, characterization,\\poignancy, and intelligence of a \textbf{bad} sitcom} & \tabincell{l}{a dark, \textbf{dull} \textit{thriller} with a parting shot\\that misfires} & \tabincell{l}{$\dots$one resurrection \textbf{too} \textit{many}}\\
\bottomrule
\end{tabular}
\caption{Example snippets with positive and negative sentiments from the RottenTomatoes dataset.
The most important word in a snippet found by \textbf{DFAX} is shown in boldface, while that found by \textit{DLIME} is in italic.
}
\label{tab:RT}
\end{table*}

\begin{figure*}[t]
\centering
\includegraphics[width=0.94\textwidth]{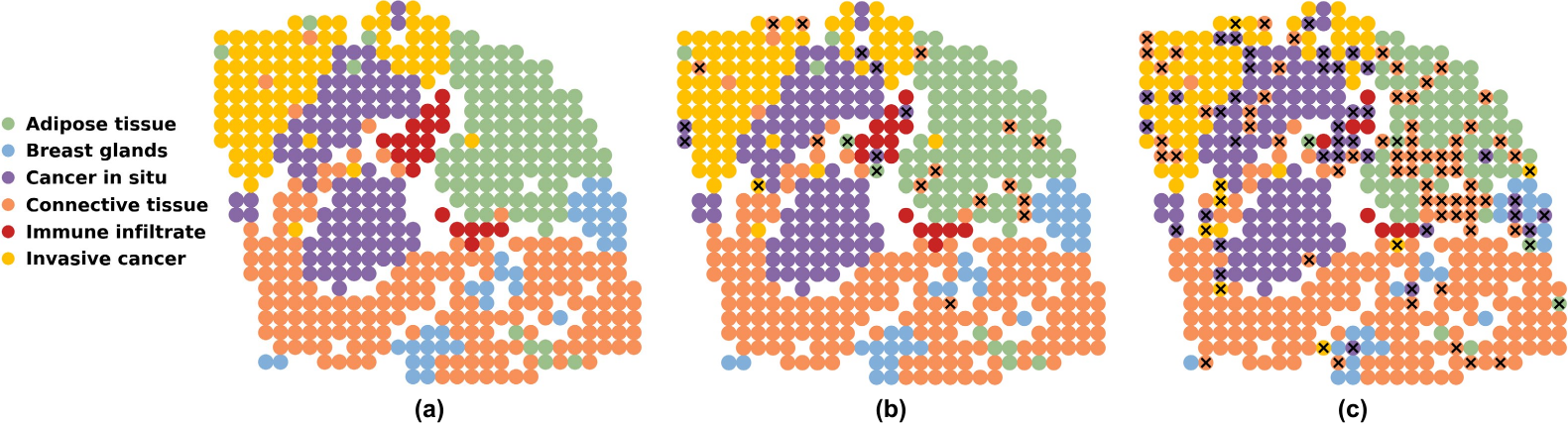}
\caption{Visualization of tissue predictions on the HER2st dataset.
(a) Initial predictions for the original cells with all 314 genes.
(b-c) Predictions based on the 157 salient genes only, identified by (b) DFAX and (c) DLIME.
}
\label{fig:HER2st}
\end{figure*}

\subsection{Qualitative Evaluation}

Here, we aim to show through several examples that DFAX produces more intuitive attributions with better quality.

\subsubsection{RottenTomatoes.}

This dataset consists of movie review snippets from the Rotten Tomatoes website.
After TF-IDF vectorizing the snippets and acquiring the trained classifier's predictions, the task is to identify the most important word in each snippet signifying its sentiment using a feature attribution method.
DLIME is selected for comparison with DFAX in this task as it is a leading baseline method in our preceding quantitative experiments.
It serves as a representative for the baseline feature attribution methods.

As presented in Table~\ref{tab:RT}, DFAX identifies words such as ``compelling'', ``fascinating'', and ``moving'' as most indicative of positive sentiment, while highlighting ``bad'', ``dull'', and ``too'' for negative sentiment.
In contrast, DLIME fails to identify reasonable words.
For instance, it selects ``real'' over ``compelling'' for a positive review and ``humor'' over ``bad'' for a negative one.
These examples suggest the superior quality of the attributions generated by DFAX.

\subsubsection{HER2st.}

Spatial transcriptomics (ST) is a pivotal technology for scientists to profile gene expression and spatial information in tissue samples \cite{marx2021method}.
A key application of ST is in oncology, where the over-expression of the human epidermal growth factor receptor 2 (HER2) gene defines the major subtypes of breast cancer.
Here we use the HER2st dataset containing HER2-positive breast tumor data collected from an ST platform.
This dataset comprises 527 cells, each characterized by a pair of spatial coordinates and the expression values of the 314 most highly variable genes.
While previously utilized in clustering  \cite{zhang2025kernel}, this dataset is also well-suited for feature attribution, an application that allows scientists to identify which genes are most critical in determining a cell’s tissue type.

We train a classifier and obtain initial tissue predictions for each cell, which are visualized in Figure~\ref{fig:HER2st}(a), plotted using the spatial coordinates provided in the dataset.
Then a feature attribution method is employed to identify the 157 (\textit{i.e.}, half the total) most salient genes for each cell.
The remaining non-salient genes are then masked with random expression values, and the classifier's predictions are re-evaluated on these masked cells.\footnote{
It is important to distinguish this evaluation protocol from the explanatory model building process.
Definition~\ref{def:FA} requires that the explanatory model be derived from the unmodified dataset $\X$, a criterion to which our method adheres.
The data modification described here is an evaluation protocol used solely to assess any feature attribution methods, after the explanatory models are built.
The modification is applied to the testing instances rather than to $\X$.
This evaluation protocol is similar to that used in the deletion and insertion scores reported in the quantitative experiments.
}
The underlying assumption is that an effective attribution method should preserve the most critical genes, thus maintaining high prediction accuracy even after masking (the initial predictions serve as the ground truth labels for this accuracy calculation).

The prediction results for DFAX and DLIME are illustrated in Figures~\ref{fig:HER2st}(b) and \ref{fig:HER2st}(c), respectively.
Cells for which the re-evaluated predictions do not match the initial predictions are marked with a cross.
As depicted, DFAX achieves a high accuracy of 95.64\%, significantly outperforming DLIME's 79.51\%.
Notably, DFAX results in only 6 misclassifications of cancer cells, compared to 44 for DLIME.
These findings strongly demonstrate the effectiveness of DFAX and highlight its potential for real-world scientific applications.

\subsubsection{MNIST and FMNIST.}

Additional qualitative evaluations on these two image datasets can be found in Appendix~\ref{app:QE}.

\subsection{Runtime Comparison}

\begin{figure}[t]
\centering
\includegraphics[width=0.86\linewidth]{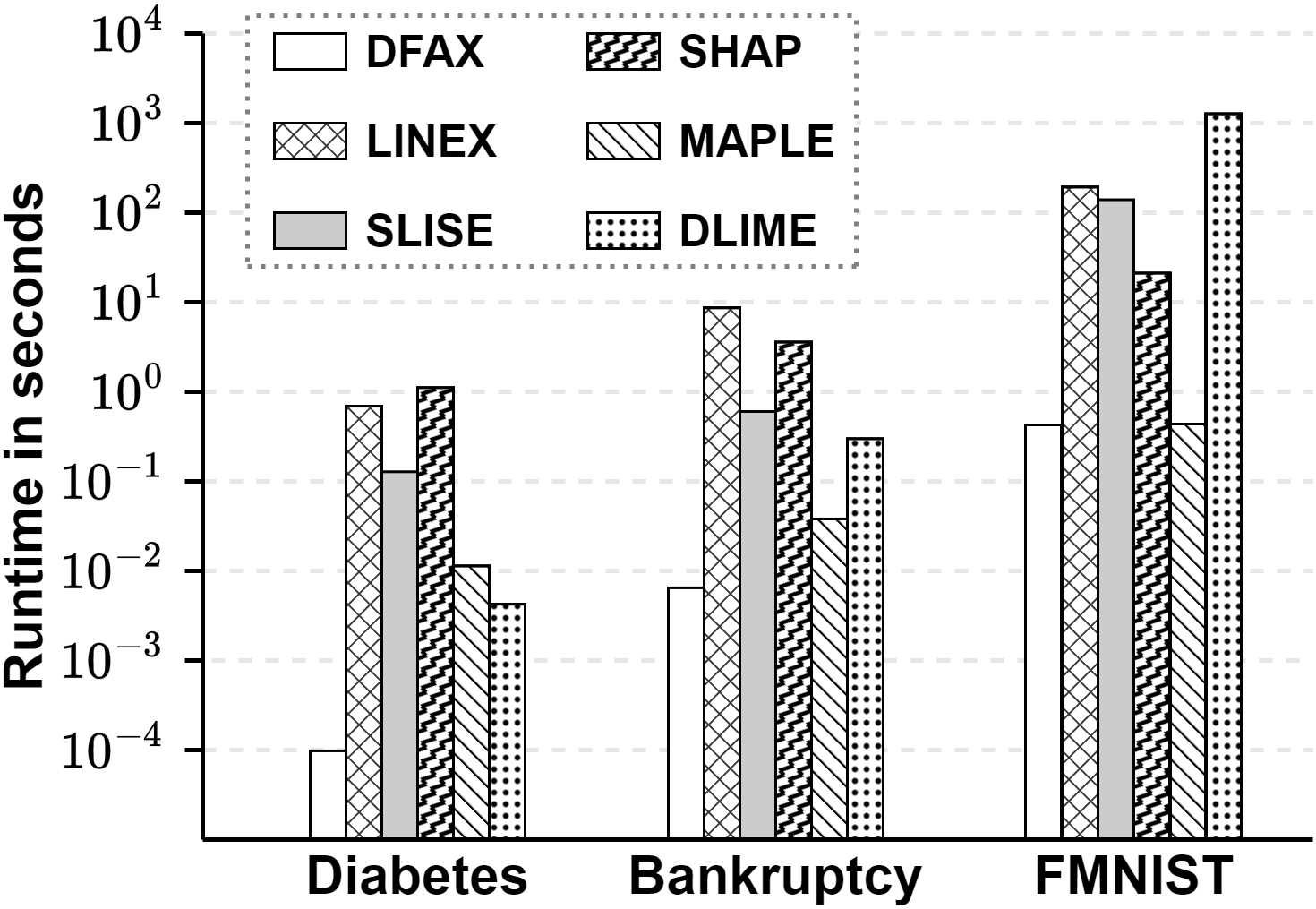}
\caption{Runtime of the methods in seconds (log-scale).}
\label{fig:time}
\end{figure}

To demonstrate the efficiency of DFAX, we provide in Figure~\ref{fig:time} an empirical runtime comparison on three representative datasets of varying sizes, feature counts, and numbers of classes.
It reports the total time required for each method to generate feature attributions for a single target instance.
Figure~\ref{fig:time} shows that DFAX achieves the shortest runtime on all three datasets.
Notably, DFAX is often faster than other baselines by orders of magnitude.
This runtime comparison underscores the efficiency of DFAX, affirming its practicality for generating feature attributions in large-scale real-world applications.

\section{Conclusions}

The proposed DFAX is the first fast and effective feature attribution explainer to operate directly from a distributional perspective by means of probability/density estimation.
This is made possible via a formal definition for the problem of feature attribution, which reveals the importance of generating explanations supported by distribution $\P$.
Our extensive experiments have demonstrated both quantitatively and qualitatively the superiority of DFAX over other state-of-the-art methods, and the efficiency of DFAX was also confirmed through runtime comparisons.

We contend that many of the limitations found in existing model-agnostic feature attribution methods stem from one underlying issue:
they have been designed and evaluated without fully understanding the problem, a direct consequence due to the absence of a formal problem definition.
The proposed formal definition addresses this gap, establishing a guideline for the evaluation and design of any explanation methods for feature attribution.

In the near future, we plan to investigate the axiomatic properties of DFAX, and to extend DFAX to the task of feature-group attribution.

\section*{Acknowledgments}

Kai Ming Ting is supported by the National Natural Science Foundation of China (Grant No. 92470116).

\bibliography{references}

\clearpage

\appendix
\setcounter{secnumdepth}{1}

\section{Hyperparameter Specifications}
\label{app:params}

The optimal hyperparameters for each method were determined via a grid search.
We selected the hyperparameter combination that yielded the best average performance in terms of the deletion and insertion scores on a validation set, which consisted of 100 instances randomly sampled from the training set.
The search spaces explored for each method's hyperparameters as well as the final hyperparameters are as follows:

\begin{itemize}
    \item DFAX$_\text{G}$ (with GKDE as the KDE):
    The only hyperparameter is $\gamma=\frac1{2\sigma^2}$ where $\sigma$ is the bandwidth of Gaussian kernel.
    
    We searched $\gamma=\{$1e-4, 1e-3, 1e-2, $\dots$, 1e4$\}$ on all the ten datasets.
    
    The final hyperparameters used are:
    $\gamma=$ 1e4 for the Diabetes dataset;
    $\gamma=$ 1e-1 for HER2st;
    $\gamma=$ 1e-2 for Rice;
    $\gamma=$ 1 for Waveform;
    $\gamma=$ 1 for Bankruptcy;
    $\gamma=$ 1e-3 for RottenTomatoes;
    $\gamma=$ 1 for Pendigits;
    $\gamma=$ 1e-2 for DryBean;
    $\gamma=$ 1e-1 for MNIST;
    and $\gamma=$ 1e-3 for FMNIST.
    
    \item DFAX$_\text{S}$ (with SiNNE as the KDE):
    The hyperparameters are the subsampling size $\psi$ and the number of ensemble models $t$.
    
    We searched $\psi=\{$2, 3, 4, $\dots$, 8$\}$ and $t=\{$200, 300, 400, $\dots$, 2000$\}$ on all the ten datasets.
    
    The final hyperparameter used are:
    $\psi=\text{2, }t=\text{1000}$ for the Diabetes dataset;
    $\psi=\text{6, }t=\text{1200}$ for HER2st;
    $\psi=\text{2, }t=\text{400}$ for Rice;
    $\psi=\text{5, }t=\text{1900}$ for Waveform;
    $\psi=\text{8, }t=\text{500}$ for Bankruptcy;
    $\psi=\text{2, }t=\text{2000}$ for RottenTomatoes;
    $\psi=\text{4, }t=\text{400}$ for Pendigits;
    $\psi=\text{5, }t=\text{2000}$ for DryBean;
    $\psi=\text{2, }t=\text{1400}$ for MNIST;
    and $\psi=\text{2, }t=\text{500}$ for FMNIST.
    
    \item LINEX:
    We implemented the random perturbation-based LINEX method according to the paper \cite{dhurandhar2023locally}.
    The hyperparameters are the kernel width $\sigma$, $l_1$ bound $t$, $l_\infty$ bound $\gamma$, the number of environments $n_\text{env}$, the number of samples $n_\text{smp}$, and the maximum number of iterations $n_\text{itr}$.
    We performed a coarse search over a wide range of hyperparameters first, followed by a fine-grained search to select the final hyperparameters.
    
    We searched $\sigma=$ [0.1, 10], $t=$ [0.01, 1], $\gamma=$ [0.01, 1], $n_\text{env}=$ [2, 5], $n_\text{smp}=$ [50, 200], $n_\text{itr}=$ [50,  200] on the Diabetes dataset;
    $\sigma=$ [1, 100], $t=$ [0.01, 1], $\gamma=$ [0.001, 0.1], $n_\text{env}=$ [2, 5], $n_\text{smp}=$ [50, 200], $n_\text{itr}=$ [50, 200] on HER2st;
    $\sigma=$ [0.001, 0.1], $t=$ [0.1, 100], $\gamma=$ [0.001, 0.1], $n_\text{env}=$ [2, 5], $n_\text{smp}=$ [100, 1000], $n_\text{itr}=$ [50, 200] on Rice;
    $\sigma=$ [1e3, 1e5], $t=$ [0.1, 100], $\gamma=$ [0.001, 0.1], $n_\text{env}=$ [2, 5], $n_\text{smp}=$ [100, 1000], $n_\text{itr}=$ [50, 200] on Waveform;
    $\sigma=$ [1e2, 1e4], $t=$ [0.1, 100], $\gamma=$ [1e-5, 1e-1], $n_\text{env}=$ [2, 5], $n_\text{smp}=$ [100, 1000], $n_\text{itr}=$ [50, 200] on Bankruptcy;
    $\sigma=$ [0.1, 100], $t=$ [0.1, 100], $\gamma=$ [0.01, 100], $n_\text{env}=$ [2, 5], $n_\text{smp}=$ [100, 1000], $n_\text{itr}=$ [50, 200] on RottenTomatoes;
    $\sigma=$ [0.1, 100], $t=$ [0.01, 10], $\gamma=$ [0.1, 10], $n_\text{env}=$ [2, 5], $n_\text{smp}=$ [100, 1000], $n_\text{itr}=$ [50, 200] on Pendigits;
    $\sigma=$ [0.01, 10], $t=$ [0.01, 10], $\gamma=$ [0.01, 1], $n_\text{env}=$ [2, 5], $n_\text{smp}=$ [100, 1000], $n_\text{itr}=$ [50, 200] on DryBean;
    $\sigma=$ [0.01, 10], $t=$ [0.01, 10], $\gamma=$ [0.01, 1], $n_\text{env}=$ [2, 5], $n_\text{smp}=$ [100, 1000], $n_\text{itr}=$ [50, 200] on MNIST;
    and $\sigma=$ [0.01, 10], $t=$ [0.01, 10], $\gamma=$ [0.001, 0.1], $n_\text{env}=$ [2, 5], $n_\text{smp}=$ [100, 1000], $n_\text{itr}=$ [50, 200] on FMNIST.

    The final hyperparameter used are:
    $\sigma=$ 1.8, $t=$ 0.1, $\gamma=$ 0.3, $n_\text{env}=$ 3, $n_\text{smp}=$ 100, $n_\text{itr}=$ 120 for the Diabetes dataset;
    $\sigma=$ 8, $t=$ 0.05, $\gamma=$ 0.004, $n_\text{env}=$ 3, $n_\text{smp}=$ 100, $n_\text{itr}=$ 100 for HER2st;
    $\sigma=$ 0.01, $t=$ 5, $\gamma=$ 0.01, $n_\text{env}=$ 2, $n_\text{smp}=$ 1000, $n_\text{itr}=$ 100 for Rice;
    $\sigma=$ 5e4, $t=$ 1, $\gamma=$ 0.01, $n_\text{env}=$ 2, $n_\text{smp}=$ 300, $n_\text{itr}=$ 100 for Waveform;
    $\sigma=$ 2e3, $t=$ 1, $\gamma=$ 1e-4, $n_\text{env}=$ 2, $n_\text{smp}=$ 300, $n_\text{itr}=$ 100 for Bankruptcy;
    $\sigma=$ 2, $t=$ 2, $\gamma=$ 0.01, $n_\text{env}=$ 3, $n_\text{smp}=$ 300, $n_\text{itr}=$ 120 for RottenTomatoes;
    $\sigma=$ 1.8, $t=$ 0.1, $\gamma=$ 0.3, $n_\text{env}=$ 3, $n_\text{smp}=$ 100, $n_\text{itr}=$ 120 for Pendigits;
    $\sigma=$ 0.1, $t=$ 0.2, $\gamma=$ 0.01, $n_\text{env}=$ 3, $n_\text{smp}=$ 200, $n_\text{itr}=$ 100 for DryBean;
    $\sigma=$ 1.8, $t=$ 0.2, $\gamma=$ 0.01, $n_\text{env}=$ 4, $n_\text{smp}=$ 250, $n_\text{itr}=$ 120 for MNIST;
    and $\sigma=$ 1.4, $t=$ 0.05, $\gamma=$ 0.001, $n_\text{env}=$ 3, $n_\text{smp}=$ 750, $n_\text{itr}=$ 120 for FMNIST.
    
    \item SLISE:
    We utilized the official Python package provided by the authors \cite{bjorklund2023explaining}.
    The hyperparameters are the error tolerance $\epsilon$, $l_1$ regularization strength $\lambda_1$, and $l_2$ regularization strength $\lambda_2$.
    We performed a two-step search as done for LINEX.

    We searched $\epsilon=$ [0.01, 1] on the Diabetes dataset;
    $\epsilon=$ [0.1, 10] on HER2st;
    $\epsilon=$ [0.01, 1] on Rice;
    $\epsilon=$ [0.001, 0.1] on Waveform;
    $\epsilon=$ [1e2, 1e4] on Bankruptcy;
    $\epsilon=$ [1e-3, 1] on RottenTomatoes;
    $\epsilon=$ [0.1, 10] on Pendigits;
    $\epsilon=$ [0.1, 10] on DryBean;
    $\epsilon=$ [1e-3, 1] on MNIST;
    and $\epsilon=$ [0.01, 1] on FMNIST.
    $\lambda_1=$ [0, 1] and $\lambda_2=$ [0, 1] were searched on all the ten datasets.

    The final hyperparameter used are:
    $\epsilon=$ 0.1, $\lambda_1=$ 0, $\lambda_2=$ 1 for the Diabetes dataset;
    $\epsilon=$ 0.02, $\lambda_1=$ 0, $\lambda_2=$ 0 for HER2st;
    $\epsilon=$ 0.01, $\lambda_1=$ 0, $\lambda_2=$ 0 for Rice;
    $\epsilon=$ 0.01, $\lambda_1=$ 0, $\lambda_2=$ 0 for Waveform;
    $\epsilon=$ 200, $\lambda_1=$ 0, $\lambda_2=$ 0 for Bankruptcy;
    $\epsilon=$ 0.01, $\lambda_1=$ 0, $\lambda_2=$ 0 for RottenTomatoes;
    $\epsilon=$ 0.4, $\lambda_1=$ 0.01, $\lambda_2=$ 0 for Pendigits;
    $\epsilon=$ 0.8, $\lambda_1=$ 0, $\lambda_2=$ 0 for DryBean;
    $\epsilon=$ 0.01, $\lambda_1=$ 0, $\lambda_2=$ 0.01 for MNIST;
    and $\epsilon=$ 0.5, $\lambda_1=$ 0, $\lambda_2=$ 0 for FMNIST.

\begin{figure*}[!t]
\centering
\includegraphics[width=\textwidth]{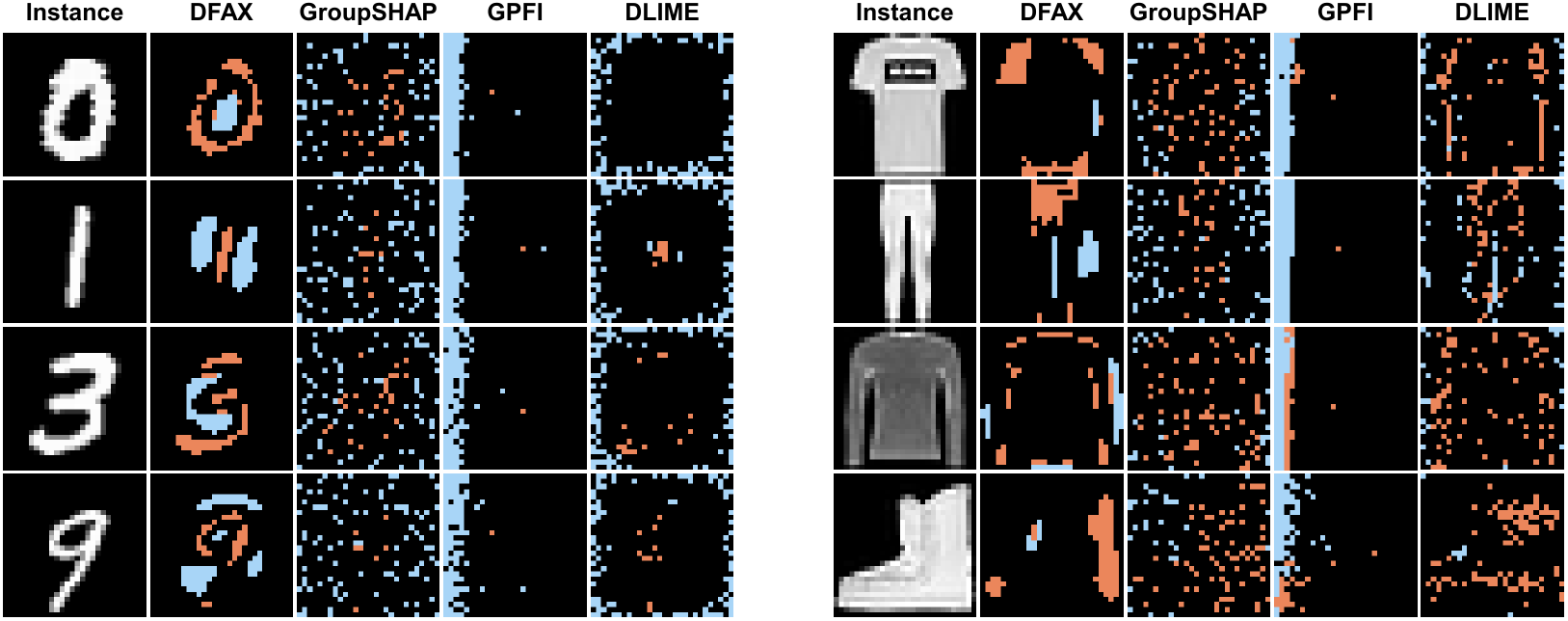}
\caption{Visualization of the 100 most salient pixels for the MNIST and FMNIST images.}
\label{fig:(F)MNIST}
\end{figure*}
    
    \item SHAP and MAPLE:
    We utilized the implementations available in the official Python library for SHAP \cite{lundberg2017unified}.
    While these methods do not require hyperparameter tuning, they necessitate a background dataset.
    Using the entire training set for this purpose is computationally infeasible, so we selected a representative background dataset of 100 samples, which represents a balance between effectiveness and efficiency.
    This size adheres to the library's recommendation, as using more than 100 background data samples triggers a warning regarding significantly slower run times.

    We explored two strategies offered by the library to select these 100 samples from the training set: random selection and k-means summarization.
    The final strategy was selected individually for each method and dataset:
    for SHAP, random selection was used on the Diabetes, Waveform, DryBean, MNIST, and FMNIST datasets, while k-means summarization was used on HER2st, Rice, Bankruptcy, RottenTomatoes, and Pendigits;
    for MAPLE, random selection was used on Waveform, Bankruptcy, Pendigits, and FMNIST, while k-means summarization was used on Diabetes, HER2st, Rice, RottenTomatoes, DryBean, and MNIST.
    
    \item DLIME:
    We used the original source code provided by its authors \cite{zafar2021deterministic}.
    This method has a single hyperparameter, the number of clusters $n_\text{c}$, for the agglomerative hierarchical clustering.

    We searched $n_\text{c}=\{$2, 3, \dots, $m\}$ on all the ten datasets, where $m$ is the total number of classes.

    The final hyperparameters used are:
    $n_\text{c}=$ 4 for the DryBean dataset;
    $n_\text{c}=$ 2 for the other nine datasets.
\end{itemize}

\section{Additional Qualitative Evaluation}
\label{app:QE}

\subsubsection{MNIST and FMNIST.}

We present here another qualitative comparison of the attributions generated by DFAX and DLIME on the MNIST and FMNIST datasets.
In addition, we include two model-agnostic methods for feature-group attribution, GroupSHAP \cite{jullum2021groupshapley} and GPFI \cite{au2022grouped}, in this comparison.
For each image instance, we identify the 100 most salient pixels based on their attribution scores.
The original images alongside the 100 most salient pixels identified by each method are visualized in Figure~\ref{fig:(F)MNIST}, where the salient pixels that overlap with the original image's foreground are shown in coral color, while those located in the vacant background areas are colored light blue.

For the MNIST images, DFAX produces insightful attributions.
Not only does it correctly identify key pixels that constitute the digits (coral), but it also highlights critical background pixels (light blue).
These background areas include the hollow center of the digit ``0'', the empty space adjacent to the ``1'', the region to the left of the ``3'', and the area within and around the ``9''.
Note that DFAX does not simply select all non-zero pixels in the original images, instead, it selectively finds those most crucial for predicting the target class, which explains the observed discontinuities in the coral pixels.
DFAX maintains such high-quality performance on the FMNIST images, where it highlights semantically meaningful features such as the shoulders of a T-shirt, the waist and the empty space between trousers, the outline of a pullover, and the shaft and heel of an ankle boot.

However, the performance of other baseline methods is substantially weaker.
While we can observe that GroupSHAP's coral pixels form the faint outlines of the digits and objects, the salient pixels overall lack clear structure and appear almost randomly distributed.
GPFI yields uninformative attributions on both datasets.
DLIME performs poorly on MNIST images, focusing mainly on outermost pixels.
Though its performance improves on FMNIST, its attributions still lack the clear semantic relevance shown by DFAX.

To summarize, DFAX demonstrates superior performance over the baseline methods, including those designed for feature-group attribution.
The visualization results confirm that DFAX effectively identifies salient features that are highly intuitive.

\end{document}